\documentclass[11pt]{article}
\usepackage[a4paper,margin=1in]{geometry}

\usepackage[utf8]{inputenc}
\usepackage[T1]{fontenc}
\usepackage{lmodern}
\usepackage[english]{babel}

\usepackage{amsmath,amssymb}

\usepackage{graphicx}
\usepackage{subcaption} 
\usepackage{caption} 

\usepackage{hyperref}
\usepackage[round,authoryear]{natbib}   
\usepackage{breakcites}                  
\usepackage{setspace}
\usepackage{url}

\title{Learning to Generate Pointing Gestures in Situated Embodied Conversational Agents\thanks{Published in Frontiers in Robotics and AI (2023). DOI: \href{https://doi.org/10.3389/frobt.2023.1110534}{10.3389/frobt.2023.1110534}}}
\date{} 

\author{Anna Deichler$^{1}$, Siyang Wang$^{1}$, Simon Alexanderson$^{1}$, Jonas Beskow$^{1}$ \\
\small $^{1}$Division of Speech, Music and Hearing, KTH Royal Institute of Technology, Stockholm, Sweden \\
\small Corresponding author: deichler@kth.se}

\begin{document}
\maketitle

\begin{abstract}

One of the main goals of robotics and intelligent agent research is to enable them to communicate with humans in physically situated settings. Human communication consists of both verbal and non-verbal modes. Recent studies in enabling communication for intelligent agents have focused on verbal modes, i.e. language and speech. However, in a situated setting the non-verbal mode is crucial for an agent to adapt flexible communication strategies. In this work, we focus on learning to generate non-verbal communicative expressions in situated embodied interactive agents. Specifically, we show that an agent can learn pointing gestures in a physically simulated environment through a combination of imitation and reinforcement learning that achieves high motion naturalness and high referential accuracy. 
We compared our proposed system against several baselines in both subjective and objective evaluations. The subjective evaluation is done in a virtual reality setting where an embodied referential game is played between the user and the agent in a shared 3D space, a setup that fully assesses the communicative capabilities of the generated gestures. The evaluations show that our model achieves a higher level of referential accuracy and motion naturalness compared to a state-of-the-art supervised learning motion synthesis model, showing the promise of our proposed system that combines imitation and reinforcement learning for generating communicative gestures. Additionally, our system is robust in a physically-simulated environment thus has the potential of being applied to robots. 

\end{abstract}
\section{Introduction}
\subsection{Overview}


Humans rely on both verbal and non-verbal modes of communications in physically situated conversational settings. In these settings, non-verbal expression, such as face and hand gestures, 
often contain information which is not present in the speech. A prominent example is how people point to an object instead of describing it with words. This complementary function makes communication more efficient and robust than speech alone. In order for embodied agents to interact with humans more effectively, they need to adapt similar strategies, i.e they need to both comprehend and generate non-verbal communicative expressions. 

Regarding comprehension, one focus has been on gesture recognition in order to accomplish multimodal reference resolution and establishing common ground between human and agent \cite{abidi2013human,haring2012studies,wu2021communicative}. Regarding generation, the main focus has been on generating co-speech gestures that typically uses supervised learning techniques to map text or speech audio to motion (see \cite{liu2021speech} for an overview). While these approaches can generate natural looking gesticulation, they only model beat gestures, a redundant aspect of non-verbal communication. Moreover, these supervised methods require large labeled datasets tied to a specific embodiment and are not physics-aware, posing a major issue for transferring the results to robots.


The focus of our study is the generation of pointing gestures. According to McNeill's influential classification, pointing gestures are among the four primary types of gestures, alongside with iconical, metaphorical and beat gestures \cite{david1992hand}. Iconic gestures depict concrete objects or actions with the hands. Metaphoric gestures express abstract concepts or ideas with the hands. Beat gestures are simple movements that emphasize or segment speech. Deictic gestures point to objects, locations, or persons in space. Pointing is the most specialized of all gestures in terms of its association with particular verbal expressions, specifically reference and referring expressions \cite{kibrik2011reference}. Linguistic reference is the act of using language to refer to entities or concepts. Linguistic elements that perform a mention of a referent are called referring expressions or referential devices. People use reference in discourse to draw the listener’s attention to a referent or target, which can be in either the speech-external (deictic reference) or speech-internal (anaphoric reference) environment. The anaphoric referent is an element of the current discourse, while the deictic referent is outside the discourse in the spatio-temporal surroundings \cite{talmy2020targeting}. Deictic expressions are used to indicate a location or point in time relative to the deictic centre, which is the center of a coordinate system that underlies the conceptualization of the speech situation. They are essential parts of human communication, since they establish a direct referential link between world and language. Demonstratives like "this", "that" or "there" are the simplest form of deictic expressions that focus the interlocutor’s attention to concrete entities in the surrounding situation \cite{peeters2021conceptual}. In everyday conversations people often use non-verbal means (e.g. eyes, head, posture, hands) to indicate the location of the referent alongside with verbal expressions to describe it. Pointing and eye gaze are the most prominent nonverbal means of deictic reference and also play a key role in establishing joint attention in human interactions \cite{diessel2020demonstratives}, a prerequisite for coordination in physical spaces. Pointing is ubiquitous in adult interaction across settings and it has been described as “a basic building block” of human communication \cite{kita2003pointing}. Demonstratives produced with pointing gestures are more basic than demonstratives produced without pointing \cite{cooperrider2020fifteen}. Speakers also tend to prefer short deictic descriptions when gesture is available, in contrast to longer nondeictic descriptions when gesture is not available \cite{bangerter2004using}. 

We proposed a new framework based on imitation and Reinforcement Learning (imitation-RL) for this for generating referential pointing gestures in physical situated environment. Our method, adapted from \cite{peng2021amp}, learns a motor control policy that imitates examples of pointing motions while pointing accurately. 
By providing only a few pointing gesture demonstrations, the model learns generalizable accurate pointing through combined naturalness and accuracy rewards. In contrast to supervised learning, our method achieves high pointing accuracy, requires only small amounts of data, learns physically valid motion and has high perceived motion naturalness. It is also considerably more lightweight than compared supervised-learning methods. To implement the method, we first collected a small dataset of motion-captured pointing gesture with accurate pointing positions recorded. We then trained a motor control policy on a humanoid character in a physically simulated environment using  our method. The generated gestures were then compared against several baselines to probe both motion naturalness and pointing accuracy. This was done through two subjective user studies in a custom-made virtual reality (VR) environment. In the accuracy test, a full-body virtual character points at one of several objects, and the user is asked to choose the object that the character pointed at, a setup known as referential game \cite{lazaridou2018emergence}, \cite{steels2001language}. This setup is also motivated by the notion that language use is a triadic behaviour involving the speaker, the hearer, and the entities talked about \cite{buhler1934sprachtheorie}.

We found that the proposed imitation-RL system indeed learns highly accurate pointing gestures while retaining high level of motion naturalness. It is also shown that imitation-RL is much better at both accuracy and naturalness than a supervised learning baseline. The results suggest that imitation-RL is a promising approach to generate communicative gestures.



\subsection{Related work}

Our work concerns learning communication in embodied agents through reinforcement learning, specifically learning pointing gestures in a physically simulated environment with imtation-RL. 

Below we first review related work in the area of nonverbal deictic expression and gesture generation for embodied agents, followed by an overview of two 
reinforcement learning fields related to our work: (a) reinforcement learning for communication, (b) reinforcement learning for motion control. We believe that our work is the first that learns non-verbal communication with reinforcement learning, which can be seen as bridging the two currently disjunct reinforcement learning fields.


\subsubsection{Gesture and nonverbal deictic expression generation}
There has been substantial work considering the generation of speech-driven gestures in embodied agents. Early work in synthesis of gestures for virtual agents employed methods based on rules, inverse kinematics and procedural animation to generate co-speech gestures \cite{cassell1994animated,cassell2001beat,kopp2004synthesizing,ng2010synchronized,marsella2013virtual}. Rule-based approaches, however, often require laborious manual tuning and often struggle to achieve a natural motion quality. More recently, supervised learning systems for co-speech gesture generation have achieved high naturalness in certain settings (\cite{liu2021speech}). Some of these systems take only speech audio as input  (\cite{hasegawa2018evaluation,kucherenko2021moving,ginosar2019learning,ferstl2020adversarial,alexanderson2020style}) but are limited to generating beat gestures aligned with speech. Extending these supervised learning based gesture generation methods to produce iconic, metaphoric or deictic gestures would require more elaborate modelling and a large dataset containing these gestures \cite{ferstl2018investigating}. Adding text as input can potentially generate more semantic gestures but is still limited \cite{kucherenko2020gesticulator,yoon2020speech,ahuja2020no,korzun2020finemotion}. Common for all these methods is that they do not take into account the surrounding environment and thus cannot communicate other than redundant information. On the contrary, works related to deictic expression generation for intelligent agents take into account the physical environment in which the agent is situated. Pointing and gaze generation in general have been  extensively studied in virtual agents and robots. Rule-based approaches for deictic expression generation are often used in virtual agents \cite{noma2000design,rickel1999animated,lester1999deictic}, as well as robotics. \cite{fang2015embodied} and \cite{sauppe2014robot} both implemented rule-based pointing gestures in a humanoid robot for referential communication. \cite{holladay2014legible} proposes a mathematical optimization approach to make legible pointing, i.e. disambiguation of closely situated objects. In \cite{sugiyama2007natural} natural generate rule-based deictic expressions consisting of pointing and a reference term are generated in an interactive setting with a robot. Referential gaze has been studied in virtual agents \cite{bailly2010gaze}, \cite{andrist2017looking}, as well as robotics \cite{mutlu2009nonverbal}. Some studies have adopted learning-based approaches to develop pointing gestures generation. \cite{huang2014learning} proposes a learning based modeling approach using dynamic Bayesian network to model speech, gaze and gesture behavior in a narration task. \cite{zabala2022towards} developed a system of automatic gesture generation for a humanoid robot that combines GAN based beat gesture generation with probabilistic semantic gesture insertion, including pointing gestures. \cite{chao2014developmental} takes a developmental robotics approach to generating pointing gestures using reinforcement learning for a 2D reaching task. None of the above works consider human-like and physics based pointing motion generation in 3D space.


\subsubsection{RL for communication in task-oriented embodied agents}
There have been a substantial amount of works in recent years that focused on extending the capabilities of embodied RL agents with language. This field has been reviewed in a recent survey \cite{luketina2019survey} which separates out these works into two categories: language-conditioned RL, where language is part of the task formulation and language-assisted RL, where language provides useful information for the agent to solve the task. 
The simplest form of language-conditioned is instruction following, in which a verbal instruction of the task is provided to the agent. Examples are manipulation \cite{stepputtis2019imitation} and navigation task \cite{qi2020reverie}. These studies focused more on comprehension of verbal expressions and usually no interaction takes place between the user and the agent. In more complex settings, the agent interacts with humans using natural language. In \cite{lynch2022interactive} the authors present a RL and imitation based framework, Interactive Language, that is capable of continuously adjusting its behavior to natural language based instructions in a real-time interactive setting. There have also been a recent wave of datasets and benchmarks created by utilizing 3D household simulators and crowd sourcing tools to collect large-scale task-oriented dialogue aimed at improving the interactive language capabilities of embodied task-oriented agents  \cite{padmakumar2022teach},\cite{gao2022dialfred}, \cite{team2021creating}. Most of the above mentioned works focus on the verbal mode of communication and largely on the comprehension side (e.g. instruction following). Less work has explored the non-verbal communication in situated embodied agents \cite{wu2021communicative}, especially on the generation side, which we aim to address in our work.

\subsubsection{Learning Motor Control through RL}
RL has been extensively applied to learning motor control for both robotics and graphics applications \cite{duan2016benchmarking,heess2017emergence}. The learned motion dynamic for humanoid characters from these methods are usually not very human-like. To improve both the motion dynamics and also to facilitate learning, it is proposed for the learning agent to imitate expert demonstration or motion-capture animation \cite{merel2017learning, ho2016generative}. However, it is not until DeepMimic \cite{peng2018deepmimic} that the RL-imitation approach achieved human-like motion. The drawback in DeepMimic is that it uses several manually tuned hyper-parameters in the imitation reward, a problem addressed by a follow-up work Adversarial Motion Prior (AMP) \cite{peng2021amp} which replaces the “hard" imitation in DeepMimic with a learned discriminator (similar to GAN). Through a combination of imitation reward and task reward, methods like AMP can learn human-like motion dynamics while completing a manually defined task, for example heading in a given direction while walking or spin-kick a given target. It has been shown that even with just one motion clip as imitation target, AMP learns generalizable motion control given random task input (e.g. randomly sampled walking direction) while maintaining high motion naturalness.
\newpage
\section{Materials and Methods}
\subsection{Pointing gesture data collection}
Our method for synthesizing human-like pointing gestures for the humanoid character requires an appropriate full-body motion capture dataset with diverse and accurate target locations covering the 3D space surrounding the character. Multiple datasets exist of referring expression with pointing as nonverbal modality 
\cite{schauerte2010focusing,matuszek2014learning,shukla2015probabilistic,chen2021yourefit}, but since most of these focus on comprehension, they lack full body motion capture, which is essential for our method. Existing full-body motion capture datasets for gesture generation \cite{ferstl2018investigating}, \cite{ferstl2020adversarial} focus on beat gestures with the aim to train speech-to-gesture generative models and do not contain pointing gestures. A recent dataset \cite{islamcaesar} records multi-modal referring expressions, including pointing gestures, but it is restricted to a tabletop setting.

To obtain a collection of ground truth examples for training and evaluation, we recorded a pointing gesture dataset in a optical motion capture studio equipped with 16 Optitrack Prime 41 cameras. The actor wore a suit with 50 passive markers, a pair of Manus data gloves (hand motion capture) and a head-mounted iPhone 12 (face motion and voice capture). In addition, we also recorded a pointing target, consisting of a rigid structure equipped with 4 markers. The marker data was solved to a representation of the actor's skeleton and the location of the target in 3D space using the systems software (Motive 3.0). 

Three different pointing tasks were recorded: single target pointing, two targets selection and two targets moving (point-and-place). In each setting the targets were moved around in order to get a coverage of the surrounding space, while the actor was in a stationary position. Between each pointing task the actor returned to a neutral stance. The beginning and the end of the movement is defined by the pointing hand leaving and returning to the initial, downward position. We used this information to parse the continuous recording of motion capture data into pointing clips. For more details, see \cite{deichler2022towards}.

\begin{figure}[t]
  \centering
  \begin{minipage}[b]{0.48\linewidth}
    \centering
    \includegraphics[width=\linewidth]{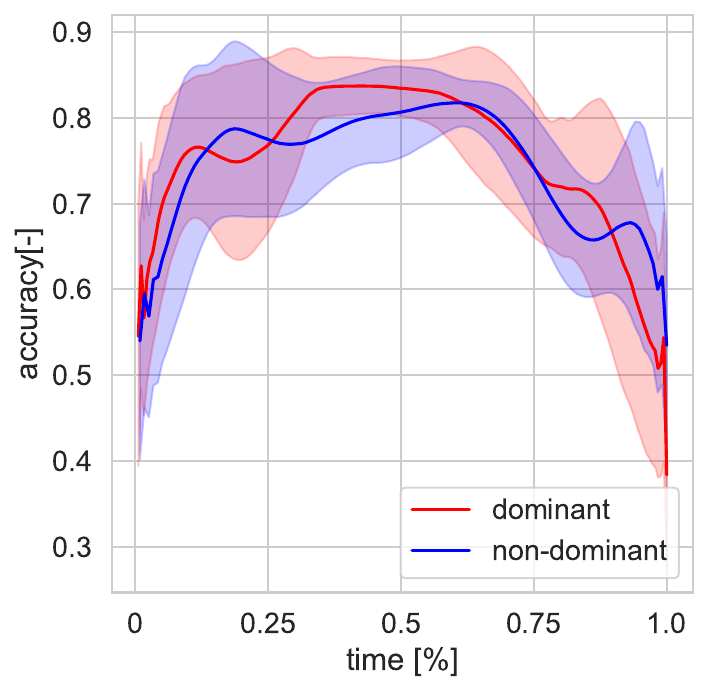}
    \caption{Averaged normalized pointing accuracy profile for front targets in training set for the dominant and non-dominant hands.}
    \label{fig:data_rew}
  \end{minipage}\hfill
  \begin{minipage}[b]{0.48\linewidth}
    \centering
    \includegraphics[width=\linewidth]{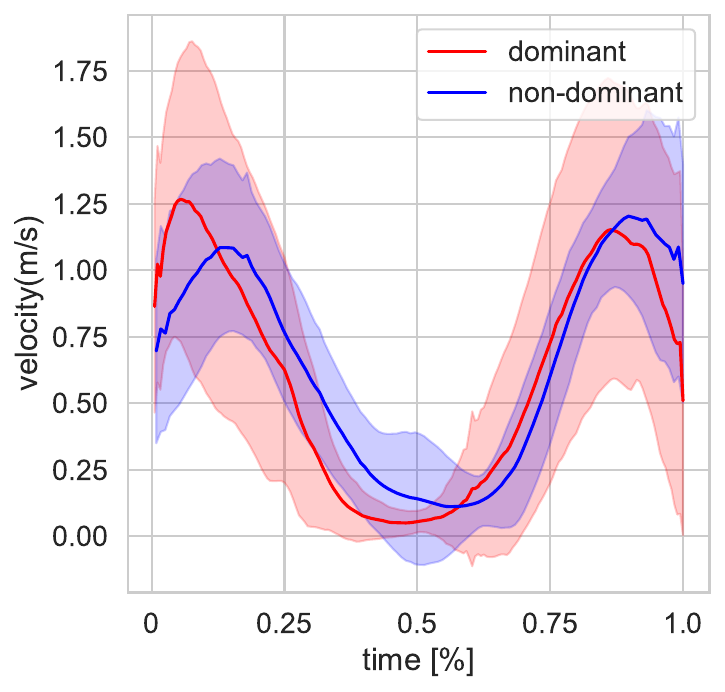}
    \caption{Averaged hand velocity profile for front targets in training set for the dominant and non-dominant hands.}
        \label{fig:data_vel}

  \end{minipage}

\end{figure}

We focus on single target pointing in this study. Here, we provide an overview of that data. We first divided the pointing target positions to front and back. We only use the front data in our study. The total sum of single target positions in the dataset is 83, from which 52 are front target pointing movements (25 left-handed and 27 right-handed). The actor’s dominant hand is right hand.  The target distribution of this subset  is visualized in 3D space in Figure \ref{fig:grid_targets} (blue balls). The single target dataset is parsed into single arm pointing movement based on peaks in displacement on the sagittal plane (yz). 
We further analyse the pointing movements in terms of pointing accuracy (Figure \ref{fig:data_rew}) and velocity (Figure \ref{fig:data_vel}) dividing into dominant (right) and non-dominant (left) hand. The pointing accuracy is calculated based on inverse of alignment angle between arm and arm to pointing target (Equation \ref{eq:pt_rew}), the more accurate pointing the higher this measure is. As seen in Figure \ref{fig:data_rew}, both dominant and non-dominant hands have a bell-shaped accuracy curve. This reflects the pointing motion trajectory: the hand first moves to pointing position as the accuracy rises, it stays there for some time (holding), and then retracts. Moreover, we note that neither hand achieves maximum accuracy. This shows that human pointing is not accurate in terms of alignment angle, which has  been shown in previous  pointing gesture  studies \cite{lucking2015pointing}. The velocity trajectory  (Figure \ref{fig:data_vel}) shows clear correspondence with accuracy trajectory. No significant difference between dominant and non-dominant hands was found.

We applied left-right mirroring on the mocap data, which doubled the total amount of data available for model training. This is mainly due to the baselines MoGlow (section \ref{sec:moglow}) and GT-nn (section \ref{sec:gt_nn}) require large amount of data to perform well. For fair comparison, we used the same mirrored data for training all systems including our proposed method even though our proposed method does not have the same data quantity requirement as the baselines.

\subsection{Method}
Our method is based on imitation-RL method AMP \cite{peng2021amp}, capable of learning complex naturalistic motion on humanoid skeletons and showing good transferrability of simulation-learned policy to real-world robots \cite{escontrela2022adversarial}, \cite{vollenweider2022advanced}. AMP is a subsequent work of Generative Adversarial Imitation learning \cite{ho2016generative}, that takes one or multiple clips of reference motion and learns a motor control policy $\pi_{\theta}$  that imitates the motion dynamics of the reference(s) through a discriminator network $D_{\phi}$. An optional task reward can be added for the specific motion task to encourage task completion in the learned policy, for example heading in a given direction while walking. The full reward function is the combination of rewards from motion imitation $r^{I}_t $ and the task reward $r^{G}_t$, weighted respectively by $\omega^{I}, \omega^{G}$,
\begin{equation}
r_t=\omega^{I} r^{I}_t + \omega^{G} r^{G}_t. 
\end{equation} 

\begin{figure}[t]
  \centering
  \begin{minipage}[b]{0.3\linewidth}
    \centering
    \includegraphics[width=\linewidth]{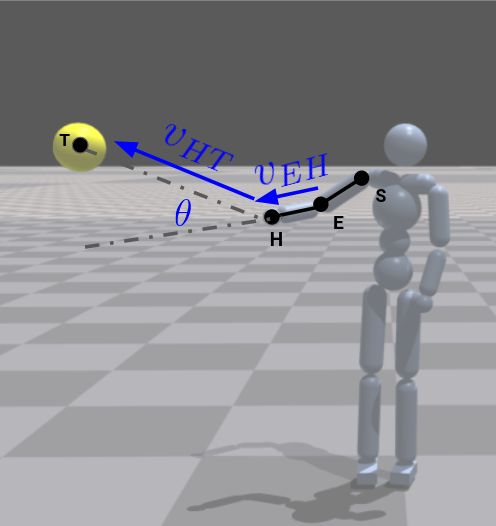}
    \caption{Pointing reward based on angle between vector from elbow joint (E) to hand joint (H) and vector from hand joint to target position (T). Agent rendered in the IsaacGym physics engine \cite{makoviychuk2021isaac}.}
    \label{fig:rew}
  \end{minipage}\hfill
  \begin{minipage}[b]{0.6\linewidth}
    \centering
    \includegraphics[width=\linewidth]{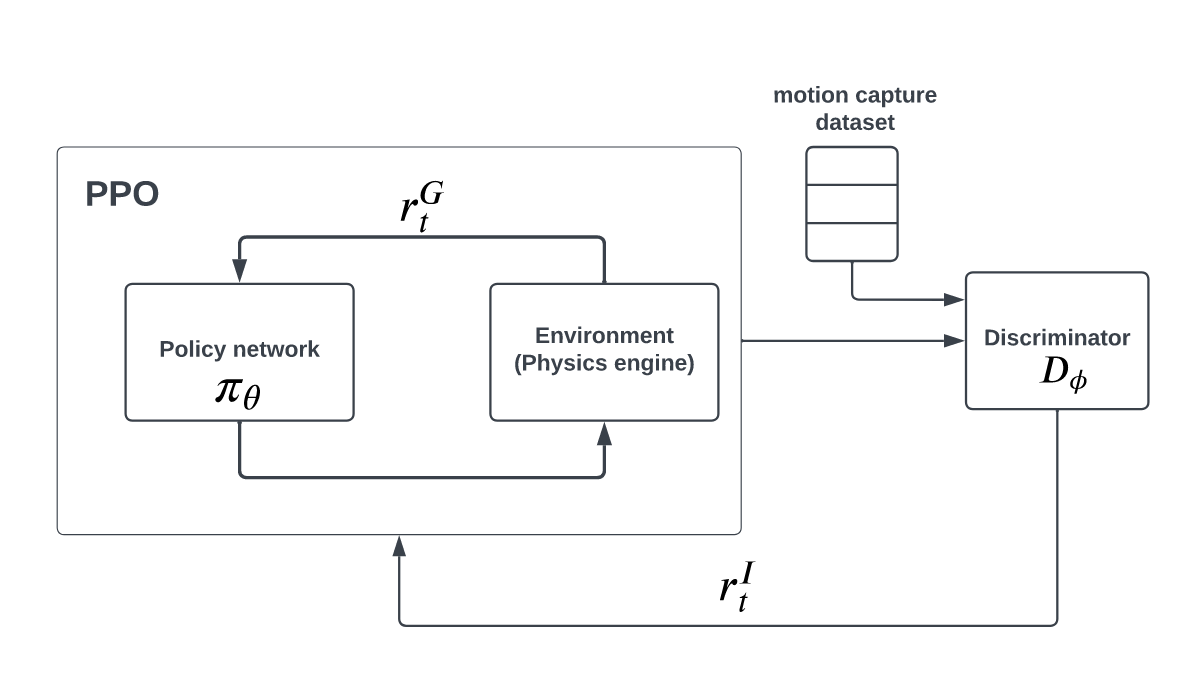}
    \caption{Schematic view of the AMP framework, which allows using a composite reward of task $r_t^{G}$ and imitation $r_t^{I}$ rewards by combining RL policy gradient methods (PPO) with a GAN-like discriminator.}
    \label{fig:amp_diagram}
  \end{minipage}

\end{figure}

The policies are trained using a combination of GAIL and PPO (Proximal Policy Optimization \cite{schulman2017proximal}). The schematic view of the system is presented on Figure \ref{fig:amp_diagram}. We kept $\omega^I=\omega^G=0.5$ and  in all AMP models. We used same action and state space as humanoid motion example presented in original AMP \cite{peng2021amp}. The state is defined as the character’s configuration in the character’s local root joint based coordinate system.  The state vector includes the relative positions of each link with respect to the root, their rotation and linear and angular velocities. The actions specify the target positions in the PD controllers for the controllable joints.

We designed a reward function $r_t^{Pt}$ for pointing accuracy 
that takes the angle between the hand-to-target vector $\overrightarrow{V_{HT}}$ and the elbow-to-hand vector $\overrightarrow{V_{EH}}$, and gives higher reward to smaller angle, i.e. more accurate pointing. This is illustrated in Figure \ref{fig:rew}. 
The angular  distance from the index finger ray has been used as a heuristics for pointing accuracy \cite{lucking2015pointing}, here we use the lower-arm extension, due to the limitations of the simulated agent's morphology. Similarly to previous closely related work in humanoid motion generation with RL and imitation \cite{peng2021amp},\cite{peng2018deepmimic}, we apply exponential scaling to angular distance term to form the task reward function as follows,
\begin{equation}
\hat{\theta} = 1- \frac{\angle (\overrightarrow{V_{HT}}, \overrightarrow{V_{EH}})}{\pi}
   \quad\text{,}\quad 
r_t^{Pt}=\frac{e^{\hat{\theta}}-1}{e}
\end{equation}\label{eq:pt_rew}

Since our task is to point at a given position, the x-y-z coordinates of the pointing position are concatenated to policy newtork input. Additionally, it is common in methods like AMP to sample the task input randomly to achieve good generalization \cite{peng2021amp}\cite{peng2018deepmimic}, for example to randomly sample walking directions when learning directional walking. We also applied this approach by randomly sampling pointing target positions in the training process. The sampling process takes in a ground-truth pointing position (corresponding to the current pointing gesture demonstration) as input and uniformly samples points within a square box of size 20x20x20 cm centered at that ground-truth position.

Through preliminary experiments, we found that the original AMP algorithm does not learn pointing gesture effectively. We subsequently implemented changes described in the following sections to improve AMP in learning pointing gesture.

\subsubsection{Phase input to network: AMP base}
\label{sec:amp_phase}
In motion imitation it is common to provide as input normalized running time called phase ($p\in [0,1]$, 0 is beginning of a clip, 1 is the end)  to the policy network in order to synchronize the simulated character with the reference motion. This was proven to be effective in imitation of single motion clips in DeepMimic \cite{peng2018deepmimic}. In AMP this is not needed, since the policies are not trained to explicitly imitate a single motion clip. However, we found that vanilla AMP without phase input struggles to learn pointing gesture motion dynamics, thus we provided the policy network with phase to help the system learn the dynamics of the pointing motion. We call this model AMP base.

\subsubsection{Phase-functioned neural network with pointing target clustering: AMP-pfnn}
\label{sec:amp_pfnn}

We found that the discriminator network  in AMP, which learns what “plausible" motion looks like through adversarial loss and provide imitation reward to the policy network, has difficulty learning pointing gesture dynamics even with phase as input. Specifically, it has trouble distinguishing between the raising phase of pointing gesture and the retraction phase, resulting in the learned motion stuck in the middle between idle and pointing positions. This is unsurprising since both phases of the pointing gesture have similar trajectories (in opposite directions). We tackled this issue by changing discriminator network from a simple MLP (as in original AMP) to a phase-functioned neural network (PFNN) \cite{holden2017phase}. 

Through early experiments, we also found that having a single network to learn pointing in all directions can be challenging. Thus, we grouped the training set pointing positions into clusters in the 3D space. The clustering is done with  the MeanShift algorithm, which is based on kernel density estimation (KDE). Using Scott's rule the estimated bandwidth is 0.78 for the univariate KDE, which results in 8 clusters in the target space. The estimated bandwidth is quite wide given the target space limits, but this is expected with a limited amount of data.  Based on this clustering, we separated the motion clips into subgroups and trained specialized networks for the different subgroups. During test time, the subgroup which contains the closest training target to the test target in terms of Euclidean distance is chosen to generate the pointing movement. We denote this version of AMP with pfnn and cluster group training as AMP-pfnn.

\subsection{Implementation}
We use a GPU-based physics simulation software Isaac gym \cite{makoviychuk2021isaac}, a rendered example of which can be seen in Figure \ref{fig:rew}. For implementation of our proposed method, we modified an existing AMP implementation provided by authors of Isaac gym \footnote{https://github.com/NVIDIA-Omniverse/IsaacGymEnvs}. The modifications also include enabling learning pointing gesture in humanoid characters and adding our proposed reward function for pointing accuracy. 

\subsection{Experiment}
\label{sec:exp}

The goal of our proposed system is to produce natural and accurate pointing gestures in an interactive agent. In order to evaluate both aspects in an embodied interaction, we created a novel virtual reality (VR) based perceptual evaluation using a 3D game engine. This is described with more details in section \ref{sec:vr_test}. We further probed the specific aspects of our proposed model in an object evaluation. We benchmarked our proposed method with two baselines, a supervised learning motion synthesis model MoGlow, and a simple nearest-neighbour retrieval method. 

\subsubsection{Baselines}
\paragraph{MoGlow}
\label{sec:moglow}

We used a state-of-the-art animation synthesis model MoGlow \cite{henter2020moglow} as supervised learning baseline. This model is based on machine learning framework normalizing flow \cite{dinh2014nice} \cite{dinh2016density} which directly estimates data probability distribution with invertible neural network. MoGlow utilizes a popular normalizing flow architecture Glow \cite{kingma2018glow} and is able to synthesize animation in a conditional or unconditional probabilistic manner, and has shown state-of-the-art performance in synthesizing locomotion \cite{henter2020moglow}, dancing \cite{valle2021transflower}, and gesture \cite{alexanderson2020style}.

We extended the conditional generation mode of MoGlow to achieve pointing gesture synthesis. We provided the model with a 4-dimensional vector as condition, consisting of pointing target position (x-y-z coordinates) and an indicator variable of whether the current frame is pointing or idling. This indicator variable differs from the AMP phase variable, as it takes binary values, whereas the phase variable is the normalized running time. This difference stems from a fundamental difference between the two models. The MoGlow model is based on  autoregressive supervised learning, and therefore has a ``sense of time", AMP is RL based and does not get the history as input, therefore the phase in used as an input.

The data preparation consists of first downsampling the mocap data to 20 frames per second, then cutting it into training clips with a sliding window of window length 6 seconds and step size 0.5 second. Such data preparation is consistent with prior studies using MoGlow for locomotion generation \cite{henter2020moglow} and gesture generation \cite{alexanderson2020style}. 

\paragraph{Nearest-neighbor Queried Ground-truth Animation (GT-NN)}
\label{sec:gt_nn}
We also created a simple baseline through nearest-neighbor querying of the dataset. Given an input pointing position, this baseline model queries the closest pointing position in the dataset and plays the corresponding pointing gesture animation. We call this baseline Ground-truth Nearest-Neighbor (GT-nn). This baseline is the topline in motion naturalness since it is playing back motion-captured animation. On the other hand, its performance in referential accuracy would be determined by how close is the input position to positions in the dataset. If the input is close to some pointing position already existing in the dataset, then this method is expected to perform well, otherwise poorly. Thus, comparing a more complex learned model with this simple baseline in referential accuracy should reveal how much the learned model is able to generalize beyond the training data to unseen pointing target positions.

\subsubsection{VR perceptual test}
\label{sec:vr_test}
\begin{figure}[t]
  \centering

  \begin{subfigure}[b]{0.30\textwidth}
    \centering
    \includegraphics[width=\linewidth]{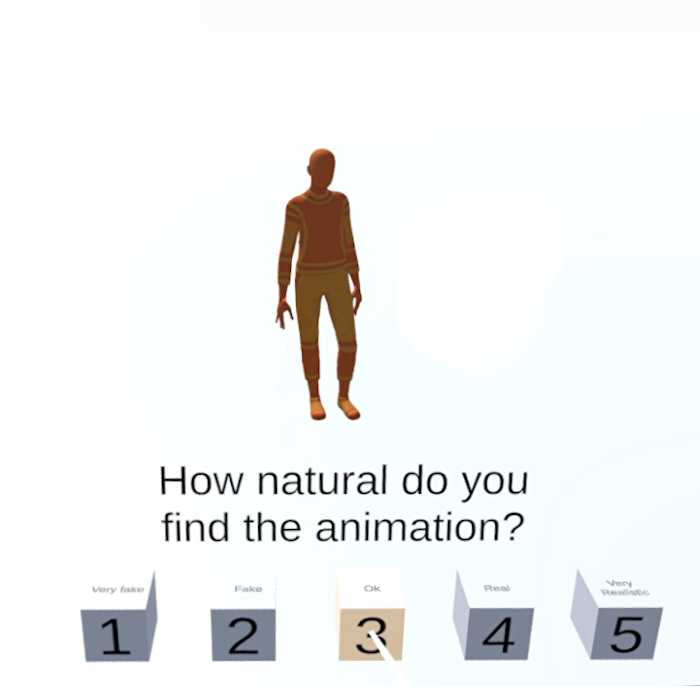}
    \caption{}
    \label{fig:VR_exp_a}
  \end{subfigure}\hfill
  \begin{subfigure}[b]{0.30\textwidth}
    \centering
    \includegraphics[width=\linewidth]{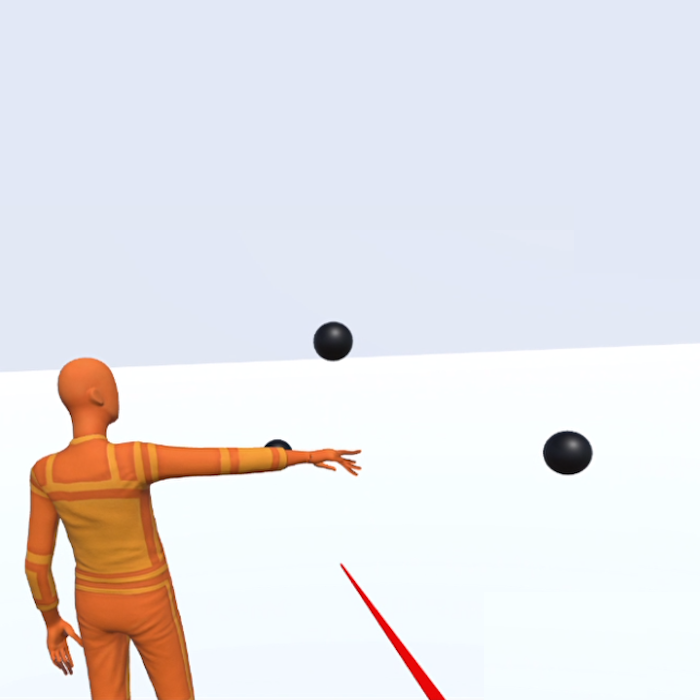}
    \caption{}
    \label{fig:VR_exp_b}
  \end{subfigure}\hfill
  \begin{subfigure}[b]{0.30\textwidth}
    \centering
    \includegraphics[width=\linewidth]{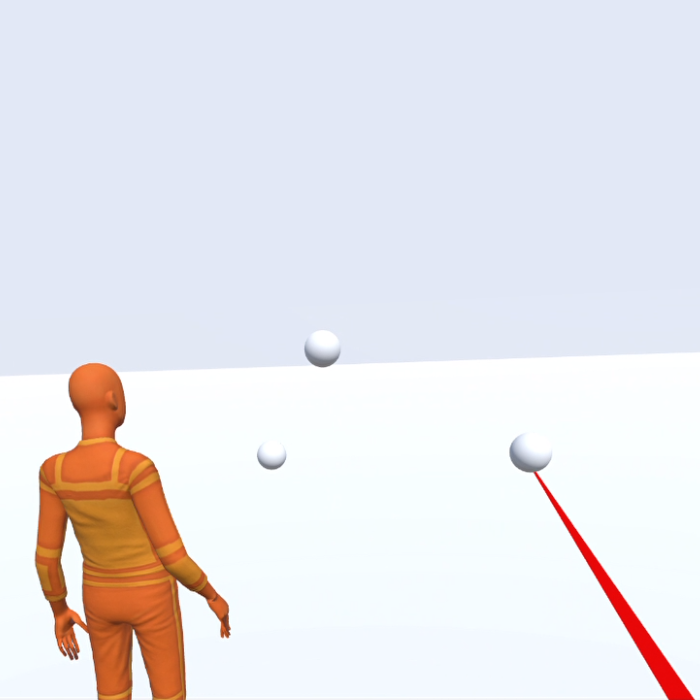}
    \caption{}
    \label{fig:VR_exp_c}
  \end{subfigure}

  \caption{Examples of user view in the VR-based perceptual test. Stage 1 (a): user rates the naturalness of the pointing gesture. Stage 2 (b,c): agent points with target and distractors (b), then the user selects the guessed object (c).}
  \label{fig:VR_exp}
\end{figure}

The perceptual test is an embodied interaction in VR. A user is put in a shared 3D space with an embodied virtual agent. In this setting, the virtual agent makes pointing gestures and the user is asked to evaluate motion naturalness of the gesture and play a referential game, i.e. guessing which object is the virtual agent pointing at. The two aspects are evaluated in two separate stages  as seen in Figure \ref{fig:VR_exp}. 

In the first stage, the naturalness of the pointing gesture is evaluated by presenting a pointing gesture to the user, then asking the user to rate “How natural do you find the animation?" on a 1-5 scale ((a) in Figure \ref{fig:VR_exp}). The avatar is in a fixed position, facing the user in this stage.  We intended for this naturalness test to only evaluate motion dynamics and not pointing accuracy, so we did not show the pointing object at this stage so that the user is not distracted by potentially inaccurate pointing. It is likely that a user would rate a pointing motion as unnatural even though the motion dynamics itself is natural but it does not point accurately at the object. We avoided this by not showing the pointing object at this stage. This evaluation setup is similar to the one in the GENEA co-speech generation challenge \cite{kucherenko2021large}. In the GENEA challenge the evaluation metrics included “human-likeness” and “appropriateness” where the first one measures motion quality, like naturalness in our study, and the second one measures task performance  - in their case how well motion matches speech, in our case the equivalent metric is pointing accuracy. 

In the second stage, the pointing accuracy is evaluated through a simplified embodied referential game \cite{steels2001language},\cite{lazaridou2018emergence}. The participants were presented with 3 balls in the environment, 1 pointing target and 2 distractors (the sampling process are described in \ref{sec:sample_test_and_distractor}). In this stage, the avatar's relative position to the user is varied between `across' and `side-by-side' conditions. In case of the `side-by-side' condition, the user stands by the shoulder of the avatar, allowing for a shared perspective on the target. In case of the `across' condition, the avatar is facing the user. After observing the agent's pointing motion, participants were asked to guess which object the agent was pointing at. The target selection mechanism was implemented using a raycast selection (red ray seen in (b)(c) of Figure \ref{fig:VR_exp}). The raycast selection becomes available to the user after the motion has ended and this is indicated to the user by the objects color changing from black to white. This setup prevents the user from selecting before seeing the entire motion. 
    
Each participant saw 5 samples for each of the 4 models in motion naturalness test (first stage) and 10 samples for each model in referential accuracy test (second stage). Thus each participant sees 20 samples in total for motion naturalness test and 40 in total for referential accuracy test. There are more samples in the referential accuracy test because a model's pointing accuracy may vary depending on the pointing position thus requires more samples to estimate true mean, while motion naturalness is easier to judge with less samples since it varies little given different pointing positions. We separated the two stages, i.e. the participant would finish motion naturalness test (first stage) before doing referential accuracy test (second stage). We also randomized ordering of the samples within each test to mitigate unwanted factors that could bias resulting statistics, such as the participant might take some time to get used to the VR game setup thus making data in the early part of the test less reliable.  Lastly, the pointing positions in both tests (5 for stage 1, 10 for stage 2) were randomly picked from the 100 sampled test positions (section \ref{sec:sample_test_and_distractor}) and were shared among models for the same user to ensure fair comparison. 

\subsubsection{Sampling test pointing targets and distractors}
\label{sec:sample_test_and_distractor}
We sampled 100 test pointing target positions within the range of ground-truth positions. The ground-truth positions roughly form a half-cylindrical shape as seen in Figure \ref{fig:grid_targets}. We thus defined the range of ground-truth position as a parameterized half-cylinder. We used 3 parameters, the height of the cylinder, the arc, and the radius of the cylinder. Given the ranges of these 3 parameters obtained from the ground-truth data, we then sampled new positions by uniformly sampling these 3 parameters within their respective ranges. The sampled 100 test positions are visualized in Figure \ref{fig:grid_targets} (red balls are sampled positions, blue balls are ground-truth).

For each sampled test position, we sampled 2 distractors in the following way. For each test position, we first sampled a distractor position the same way as the test positions (within the parameter ranges), we then checked if that distractor is within 20-40cm away from the input test position. If not, we resampled. We repeated this process until we got 2 distractors that met the distance condition. This process was done for each of the 100 test pointing positions.

We note that the distance range 20-40cm effectively determines the difficulty of the distractors. Referential game is more difficult if the distractors are closer to the actual pointing target. Through preliminary experiments, we found that 20-40cm range is difficult enough that the model would have to point with a high level of accuracy, and at the same time not too difficult that even ground-truth pointing would not be able to distinguish the correct position from distractors. Admittedly, our process is not fully theoretically driven, and future studies could explore varying distractor distance or choose this distance in a more theoretical manner.

\section{Results}

\subsection{VR perceptual test results}

\begin{figure}
\begin{center}
\includegraphics[width=\linewidth]{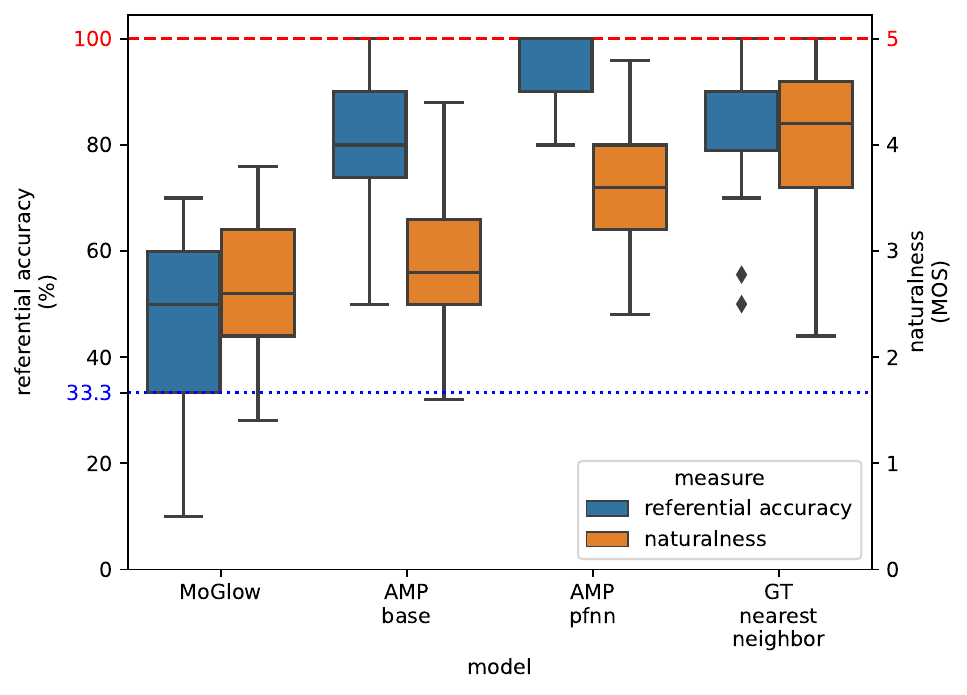}
\end{center}
\caption{VR perceptual evaluation results. Red dashed line at the top indicates maximum possible values for referential accuracy and naturalness MOS. The blue dotted line at accuracy 33.3\% corresponds to uniform guessing among 3 possible choices in the referential accuracy evaluation. All between-model differences in the two measures are significant (Wilcoxon signed-rank test with Holm-Bonferroni correction at $\alpha=0.01$), except for referential accuracy between AMP-base and GT-nn. }\label{fig:acc_nat}
\end{figure}

We recruited 39 participants (age: min=20, median=23, max=39; female: 16, male: 23). Before conducting statistical analysis on the results, we first averaged each participant's scores for each model in the two measures, referential accuracy and motion naturalness, thus obtaining a participant's average scores for the four models in the two measures. Subsequent analysis is done on this data.

Results from VR perceptual evaluation are shown in Figure \ref{fig:acc_nat}. AMP-pfnn (\ref{sec:amp_pfnn}) obtains highest referential accuracy (mean=95.5\%) and highest naturalness MOS (mean=3.57) among models. GT-nn (\ref{sec:gt_nn}) is second best overall at accuracy (mean=85.6\%) and is the naturalness MOS topline (mean=4.01) since it is playing back motion-captured animation. AMP-base (\ref{sec:amp_phase}) obtains comparable referential accuracy (mean=83.0\%) as GT-nn, but has much worse naturalness (mean=2.93) than either AMP-pfnn or topline GT-nn. MoGlow (\ref{sec:moglow}) performs worst in both accuracy (mean=45.9\%) and naturalness (mean=2.61). All but one between-sample differences are significant at $\alpha=0.01$ according to Wilcoxon signed-rank test with Holm-Bonferroni correction. The only insignificant difference is between AMP-base and GT-nn in accuracy measure. These results show that AMP-pfnn is able to point with very high referential accuracy in the presence of challenging distractors, and with high motion naturalness approaching ground-truth motion capture. Analysis on the avatar's relative position to the user in stage two shows that there is no significant between-sample  difference for the `across' and `side-by-side' conditions at $\alpha=0.01$ according to Wilcoxon signed-rank test with Holm-Bonferroni correction for any of the models. Further results on the position analysis can be found in Table \ref{tab:acc_pos_agent}.



\begin{table}[htbp]
\centering
\begin{tabular}{l||c c}
\hline
\textbf{Model} & \textbf{Across} & \textbf{Side-by-side} \\
\hline
All                 & $0.78 \pm 0.41$ & $0.76 \pm 0.42$ \\
GT nearest neighbor & $0.84 \pm 0.37$ & $0.88 \pm 0.33$ \\
MoGlow              & $0.50 \pm 0.50$ & $0.39 \pm 0.49$ \\
AMP base            & $0.83 \pm 0.38$ & $0.84 \pm 0.37$ \\
AMP PFNN            & $0.96 \pm 0.20$ & $0.95 \pm 0.22$ \\
\hline
\end{tabular}
\caption{Referential accuracy results (mean $\pm$ standard deviation) for \textit{across} and \textit{side-by-side} agent positions in VR perceptual evaluation.}
\label{tab:acc_pos_agent}
\end{table}




\subsection{Objective evaluation results}
We also conducted objective evaluations in order to gain further insights into the performance of the different models.  This is done by calculating the pointing accuracy (Equation \ref{eq:pt_rew}) over a uniform grid in the 3D space. This is also a more principled way to measure the generalization capabilities of the learning systems. In this evaluation, the pointing accuracy was calculated from the pointing reward function (Figure \ref{fig:rew}). The grid was created by taking the x,y,z limits of the training set targets and generating a spherical grid of 1000 test points (Figure \ref{fig:grid_kde}). We defined the accuracy of the pointing movement as the maximum of the reward curve \ref{fig:data_rew}, where the velocity is below 0.5 m/s in a window of frames corresponding to 0.15s around the time of pointing. Apart from the systems in the perceptual evaluation, we also added ablations to examine the effect of run time (phase) input to the AMP policy network (AMP base no runtime, compared to AMP base which has run time input \ref{sec:amp_phase}), as well as varying the number of clusters in AMP-pfnn training (\ref{sec:amp_pfnn}). Cluster variations include 1, 3 and 8, denoted as AMP-pfnn 1, AMP-pfnn 3 and AMP-pfnn 8, where the 8-cluster model is AMP-pfnn model evaluated in the perceptual test. 

\begin{figure}[ht]
  \centering

  \begin{minipage}[b]{0.45\linewidth}
    \centering
    \includegraphics[width=\linewidth]{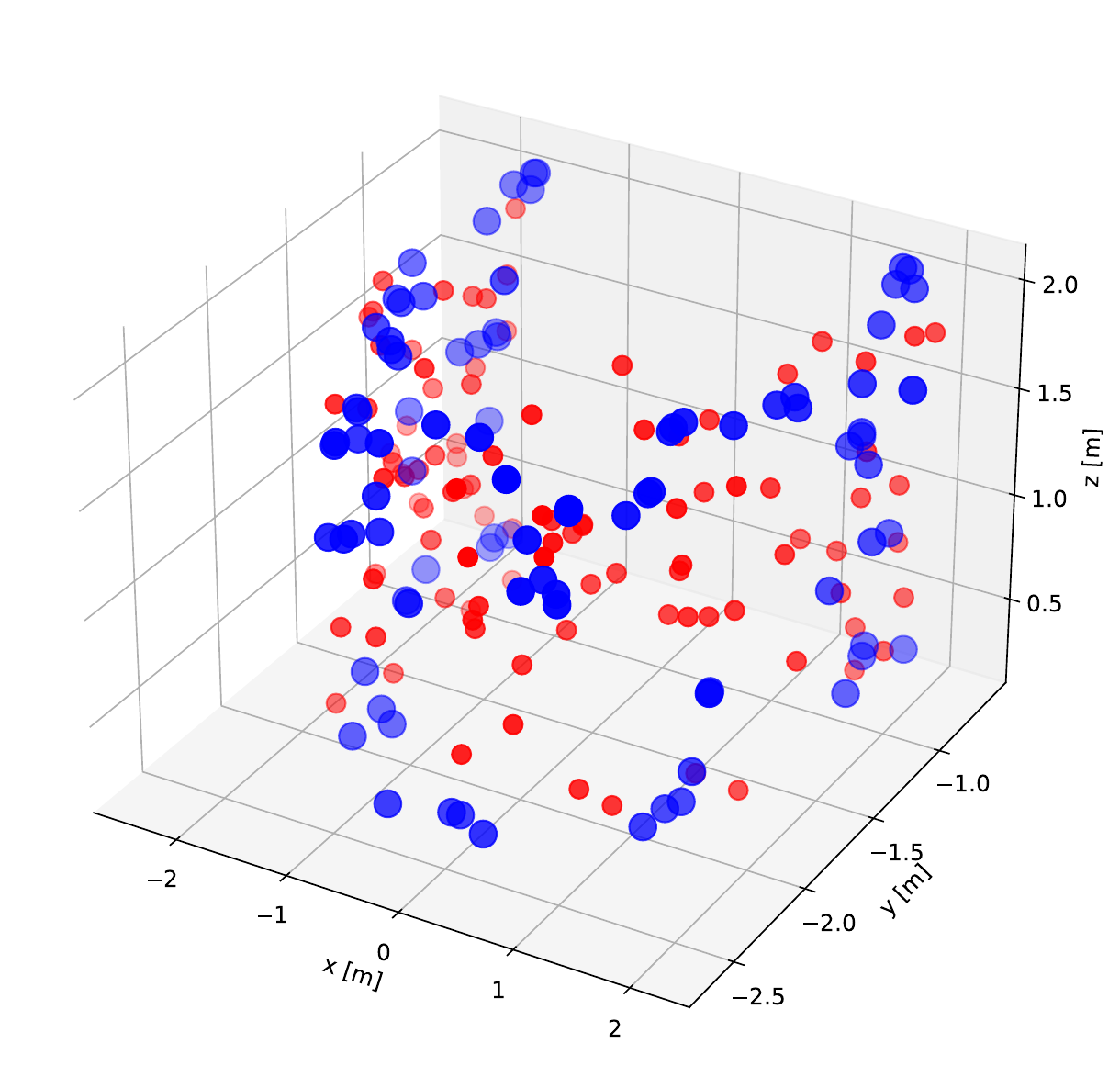}
    \caption{Targets from training set (blue) and perceptual test set (red) visualized in 3D space.}
    \label{fig:grid_targets}
  \end{minipage}
  \hfill
  \begin{minipage}[b]{0.49\linewidth}
    \centering
    \includegraphics[width=\linewidth]{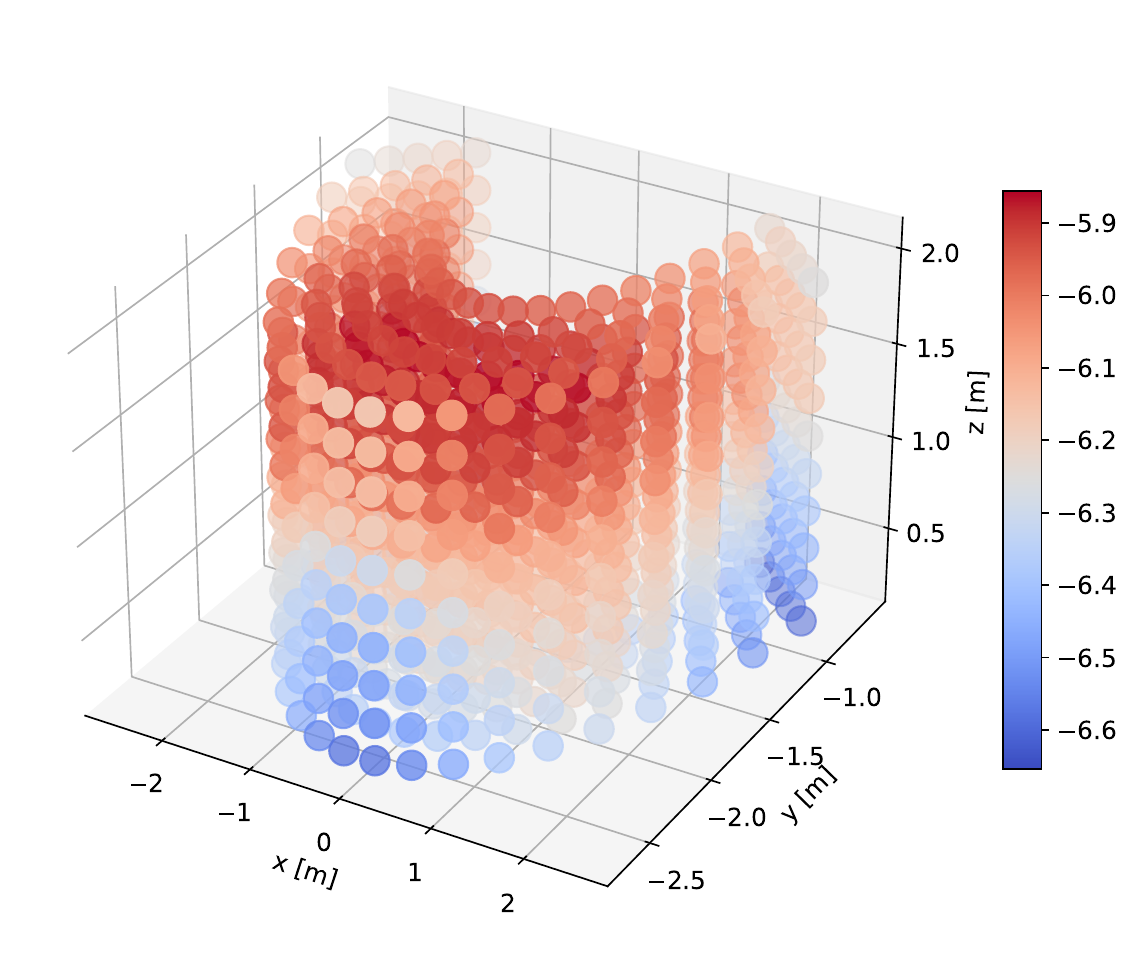}
    \caption{KDE estimation values over the 3D grid for objective evaluation (KDE fitted on training-set targets).}
    \label{fig:grid_kde}
  \end{minipage}
\end{figure}

\begin{figure}[!ht]
\begin{center}
\includegraphics[width=\linewidth]{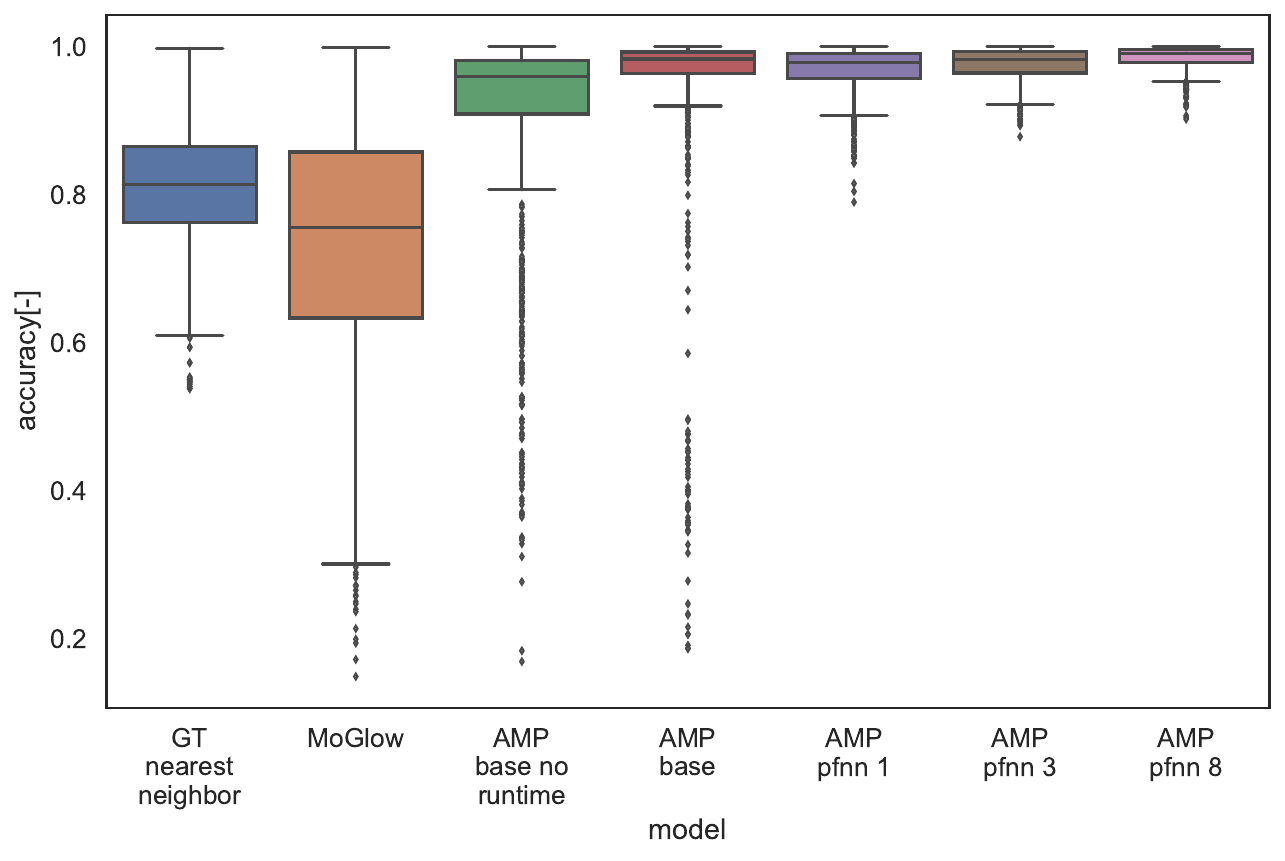}
\end{center}
\caption{Comparison of objective evaluation results of normalized pointing accuracy for the different systems. }\label{fig:acc_obj}
\end{figure}

We visualized the results for the objective evaluation in Figure \ref{fig:acc_obj}. AMP-pfnn 8 obtains the highest accuracy overall (mean=0.62), consistent with perceptual evaluation results. The graphs also shows that using 8 clusters slightly outperforms using 1 or 3 clusters and also has less outliers. This further validates the use of clusters (and specifically 8) in AMP-pfnn for learning generalizable accurate pointing. The performance gain from concatenating the run time to the policy network input is clearly visible by comparing `AMP base no runtime' (mean=0.56) and `AMP base' (mean=0.59), where the latter has the runtime input.  We can also observe that similarly to the perceptual evaluation, MoGlow performs worst amongst the compared systems (mean=0.46), again showing the limitation of a supervised system for learning communicative gesture. Overall, the results from the objective evaluation show the same performance ranking as the perceptual referential accuracy with actual users, suggesting that our objective evaluation could also be used as a system development tool for pointing gestures in the future.

Since both  MoGlow and AMP are systems learned from data, it is interesting to examine how well the models generalize outside training data distribution. We hypothesized that generalization (as measured by pointing accuracy) is correlated with how far away a test pointing position is from training position distribution. We first quantify training position distribution by fitting a multivariate kernel density estimation (KDE) on the training target positions. The bandwidth for KDE is estimated with Scott's rule and the resulting bandwidth is [0.80,0.36,0.35]. The estimated KDE density function is evaluated over 1000 test grid points as depicted on Figure \ref{fig:grid_kde}. The resulting density estimate is then correlated with the corresponding accuracy values for these points in the examined systems. Table \ref{tab:corr_acc_kde} shows the results of the correlation analysis. Only MoGlow shows strong correlation between accuracy and data density. We fully visualized the MoGlow data points in Figure \ref{fig:correlation_a}, which shows clearly that MoGlow is more accurate in test points with high training data density. We also plotted the KDE in flattened 2D heatmap (Figure \ref{fig:correlation_b}) and MoGlow's accuracy (Figure \ref{fig:correlation_c}), and it is shown that the regions with high training data density (red regions in Figure \ref{fig:correlation_b}) correspond to regions where MoGlow obtained high accuracy (red regions in Figure \ref{fig:correlation_c}). This result is not surprising since MoGlow is a supervised learning system that highly depends on training data distribution. In other systems, GT-nn, AMP base, AMP-pfnn 3, and AMP-pfnn 8 have small amount of significant correlation between training position density and model accuracy. Curiously, some of these correlations are negative, meaning that the models are more accurate further away from the training position distribution. This could be due to several factors, such as, KDE may not fully represent data density, or that the pointing position sampling mechanism in AMP models helps generalization outside of training position distribution. 

\begin{figure}[ht]
  \centering

  \begin{subfigure}[b]{0.48\textwidth}
    \centering
    \includegraphics[width=\linewidth]{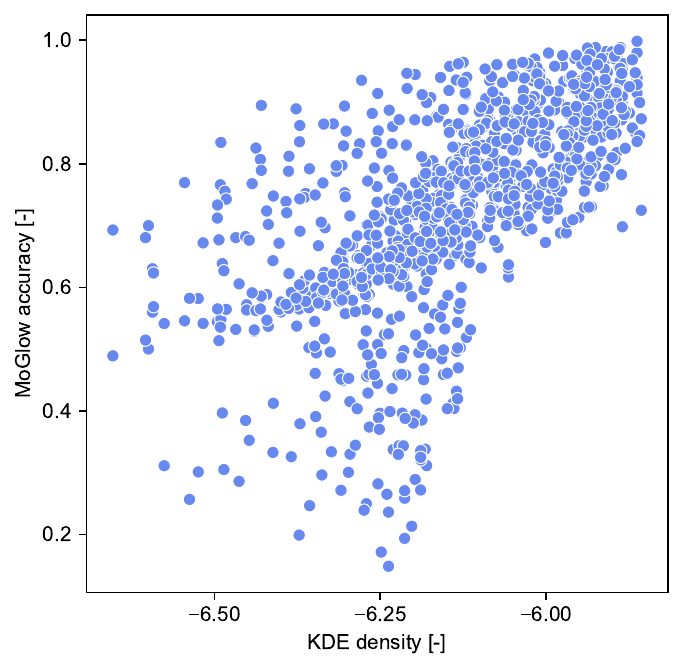}
    \caption{}
    \label{fig:grid_targets}
  \end{subfigure}%
  \hfill
  \begin{subfigure}[b]{0.48\textwidth}
    \centering
    \begin{minipage}[t]{\linewidth}
      \centering
      \includegraphics[width=\linewidth]{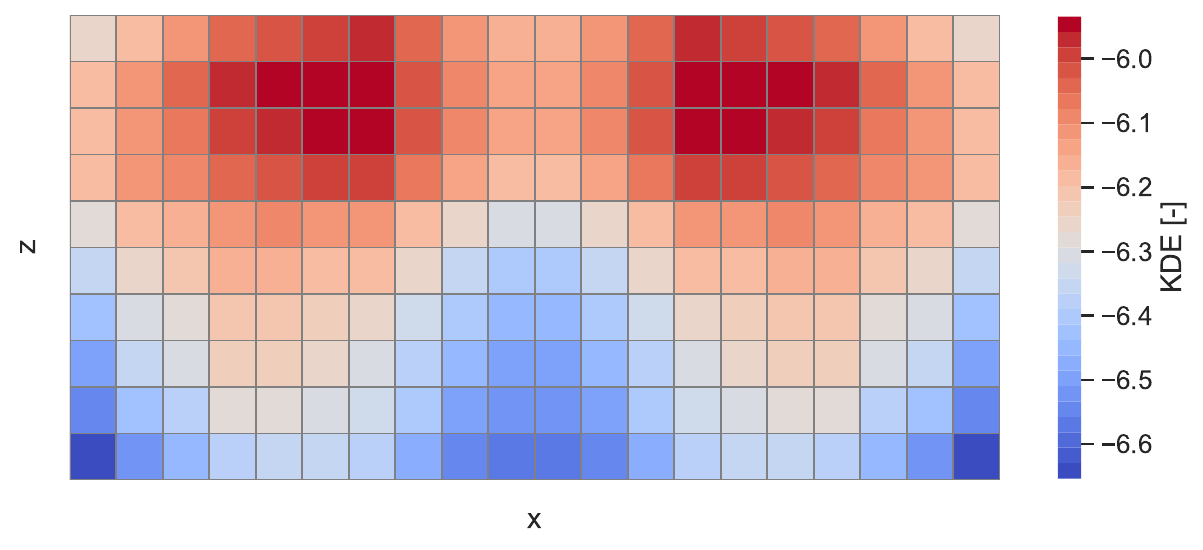}
      \caption{}
      \label{fig:grid_kde}
    \end{minipage}

    \vspace{0.8em} 

    \begin{minipage}[t]{\linewidth}
      \centering
      \includegraphics[width=\linewidth]{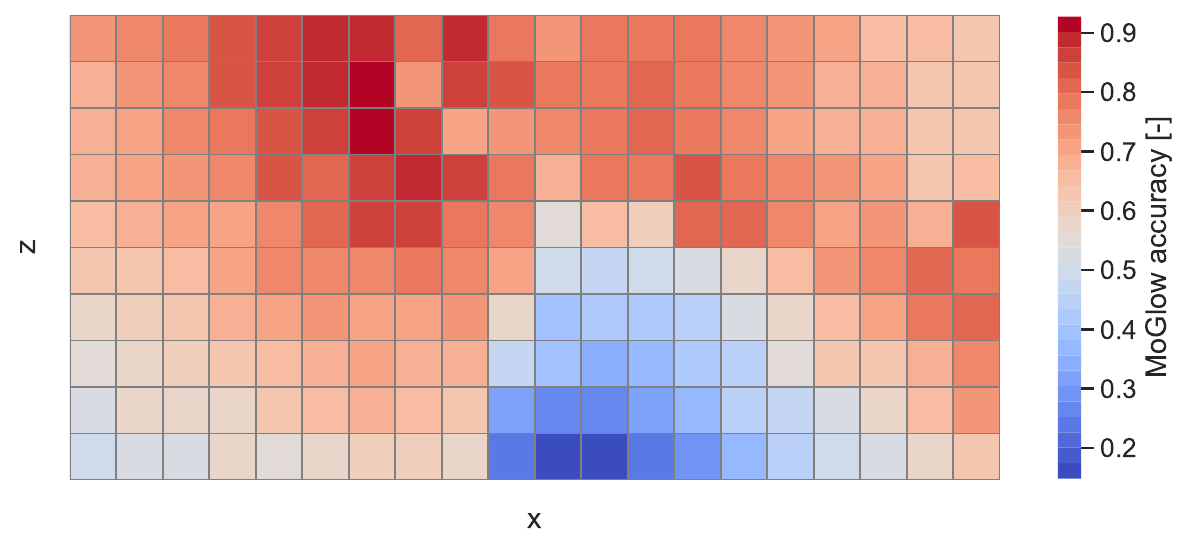}
      \caption{}
      \label{fig:grid_extra}
    \end{minipage}
  \end{subfigure}

 \caption{Correlation analysis between MoGlow model accuracy and training data density (measured by KDE). \textbf{(a)} shows moderate correlation between the accuracy and the KDE that corresponds to correlation Spearman-r=0.718 (p$<$ 0.001) as presented in Table \ref{tab:corr_acc_kde}. \textbf{(b)} and \textbf{(c)} show the KDE grid density and MoGlow accuracy respectively condensed to the x-z axis by averaging over y coordinates.}
    \label{fig:correlation}
\end{figure}
\begin{table}[htbp]
\begin{center}
\begin{tabular}{c||r    r}
\hline
\
\textbf{Model}& \textbf{Acc corr. KDE} & \textbf{p-value} \\
\hline
GT nearest neighbor & -0.118 & $<$ 0.001  \\
\hline
MoGlow & 0.718 & $<$ 0.001  \\
\hline
AMP base no runtime & -0.020 & 0.532 \\
\hline
AMP base & -0.128 & $<$ 0.001 \\
\hline
AMP-pfnn 1 & -0.043 & 0.176 \\
\hline
AMP-pfnn 3 & 0.149 & $<$ 0.001 \\
\hline
AMP-pfnn 8 & -0.135 & $<$ 0.001 \\
\hline
\end{tabular}
\caption{Correlation (Spearman-r) between accuracy and KDE density measure. }
\end{center}
\end{table}\label{tab:corr_acc_kde}


\section{Discussion}
While perceptual test results show a trend of correlation between motion naturalness and referential accuracy, i.e. models with higher naturalness also obtained higher accuracy, it is still not clear what role does motion dynamics play in referential accuracy. We did not show referential target in motion naturalness test in order to not distract naturalness judgement with some models' inaccurate pointing. But we hypothesize that the motion dynamics itself contributes to referential accuracy. A factor of motion dynamics that we already noticed could have contributed to referential accuracy is how long was the pointing phase held and how much the hand moved during that phase, i.e. how stable was the pointing phase. One of the baselines, MoGlow, sometimes has long and unstable pointing phase; it points to one position and slowly moves to a nearby position before retraction. Different users reported both choosing the initial position or the ending position as the perceived referent.  

However, one user mentioned that this type of unstable and confusing pointing gesture feels quite natural and human-like, as if the avatar is trying to convey that the pointing is not certain. This example suggests a level of communicative capacity in motion style. That is, pointing gesture style, i.e. pointing at the same position with different dynamics, can convey more information to the interlocutor than just what is the referent. It can also convey, for example, degrees of certainty of the made reference. A quick motion with short but steady pointing phase shows certainty. On the other hand, a slow motion with unsteady/constantly moving pointing phase can convey uncertainty. Previous research in cognitive science has shown that speakers tailor the kinematics of their pointing gestures to the communicative needs of the listener, by modulating the speed and duration of the different sub-components of the pointing gesture \cite{peeters2015social} and gesture kinematics has also been exploited in human robot interaction e.g. to signal hesitation \cite{moon2021design}. Future research could explore ways to model pointing styles and uncover what communicative capacity motion style has in pointing gesture.

In our current study we focused solely on the generation of nonverbal mode for referent localization in the 3D space. However, in most cases pointing gestures appear as part of multi-modal expressions, where the non-verbal mode (gesture) and verbal (language, speech)  mode carry complementary information. In these multi-modal expressions pointing gestures are accompanied by verbal descriptors, which can range from simple demonstratives (“this",“that") \cite{peeters2016and} to more complex forms  of multi-modal referring expressions describing the referent \cite{clark2004changing} (eg “that green book on the table, next to the lamp"). Furthermore, since the goal of our study was to create accurate and human-like pointing gestures, we used a simple 3D position  target representation as input to the network. In order to make the pointing gesture generation more realistic and flexible, image features could be provided to the agent. This raises further questions about efficient visual  representation for pointing and establishing eye-body coordination in embodied intelligent agents \cite{yang2021learning}.

It should also be noted that the AMP systems achieved higher pointing accuracy (Figure \ref{fig:acc_obj}) than the ground-truth training data (Figure \ref{fig:data_rew}) in terms of the defined pointing reward function \ref{eq:pt_rew}. On one hand, this is not surprising, since AMP is trained to maximize the pointing reward, as part of its full reward. On the other hand, it raises the question of how well-suited is the designed reward function for the pointing task. Pointing is an interactive communication process, which can be modeled  as a referential  game. Referential games are a type of signaling game \cite{Lewis1969-LEWCAP-4}, where the sender (gesture producer) sends a signal to the listener (observer), who needs to discover the communicative intent (localize referent) from the message.  In recent years language games have been used in emergent communication studies, mostly in text and image based agents \cite{lazaridou2018emergence}, but also in embodied agents \cite{bullard2020exploring}. 
In these, the sender and the listener jointly learn a communication protocol, often through an RL setup. This could also be relevant for our framework, since modeling the observer as an agent could substitute the current reward function based on simple geometric alignment. However, it is important to note that this kind of reward signal in emergent communication setup is very sparse, compared to our continuously defined dense reward function, which could pose further learning difficulties. Furthermore, current studies within emergent communication have largely focused on the theoretical aspects and the learned communication protocols are not interpretable to humans by design, therefore they have limited use in interactive applications. Extending these emergent communication frameworks with our pointing gesture generation framework is a promising direction in learning more general non-verbal communication in embodied interactive agent.

\section{Conclusions}
In this paper, we presented the  results of learning pointing gestures in a physically simulated embodied agent using imitation and reinforcement learning.  We conducted perceptual and objective evaluations of our proposed imitation-RL based method, and showed that it can produce highly natural pointing gestures with high referential accuracy. Moreover, we showed that our approach generalizes better in pointing gesture generation than a state-of-the-art supervised gesture synthesis model. We also presented a novel interactive VR-based method to evaluate pointing gestures in situated embodied interaction.  

\bibliographystyle{apalike} 
\bibliography{refs}    
\end{document}